\documentclass[fleqn,10pt]{wlscirep}

\usepackage{cite}

\usepackage{amsmath,amssymb,amsfonts}
\usepackage{algorithmic}
\usepackage{textcomp}
\def\BibTeX{{\rm B\kern-.05em{\sc i\kern-.025em b}\kern-.08em
    T\kern-.1667em\lower.7ex\hbox{E}\kern-.125emX}}

\newcommand{\blue}[1]{\textcolor{blue}{#1}}
\usepackage{xtab}
\usepackage{graphicx}
\usepackage{array} 
\usepackage{multirow} 
\usepackage{placeins} 
\usepackage{booktabs}    
\usepackage{ragged2e}
\usepackage{makecell}    
\usepackage{array}   
\pdfoutput=1
\usepackage{tabularx}
\usepackage{url}

\usepackage{placeins} 
\usepackage{float}     

\definecolor{Gray}{gray}{0.9}
\definecolor{darkyellow}{RGB}{250, 216, 78} 
\definecolor{salmon}{rgb}{1.0, 0.55, 0.41}
\definecolor{darkgreen}{RGB}{184, 201, 180}
\definecolor{lightgreen}{RGB}{209, 233, 201}
\definecolor{lightblue}{RGB}{211, 224, 238}
\definecolor{darkblue}{RGB}{186, 192, 217}
\definecolor{lightred}{RGB}{255,211,222}
\definecolor{darkred}{RGB}{252,150,167}
\definecolor{MagLight}{rgb}{1, 0.89, 0.8}
\usepackage{enumitem}
\usepackage{tikz}

\usepackage{tcolorbox}

  \makeatletter
\newcommand\notsotiny{\@setfontsize\notsotiny\@vipt\@viipt}
\makeatother

\usepackage[font=footnotesize]{caption}
\newcommand{\semihuge}{\fontsize{18pt}{24pt}\selectfont}

\vspace{-4mm}
\begin{abstract}
Extended reality (XR) systems, encompassing virtual reality (VR), augmented reality (AR), and mixed reality (MR), offer a transformative interface for immersive, multi-modal, and embodied human-computer interaction. 
In this paper, we envision that multi-modal multi-task (M3T) federated foundation models (FedFMs) can offer transformative capabilities for XR systems through integrating the representational strength of M3T foundation models (FMs) with the privacy-preserving and personalized model training principles of federated learning (FL). 
To this end, we first present the modular architecture of FedFMs, which entails different coordination paradigms for model training and aggregations. Afterward, we codify XR challenges that affect the implementation of FedFMs under the \textbf{SHIFT} dimensions: (1) \underline{\textbf{S}}ensor and modality diversity, (2) \underline{\textbf{H}}ardware heterogeneity and system-level constraints, (3) \underline{\textbf{I}}nteractivity and embodied personalization, (4) \underline{\textbf{F}}unctional/task variability, and (5) \underline{\textbf{T}}emporality and environmental variability. We then illustrate the manifestation of these dimensions across a set of emerging and anticipated applications of XR systems. Finally, we propose evaluation metrics, dataset requirements, and design tradeoffs necessary for the development of resource-efficient FedFMs in XR ecosystems.
\end{abstract}

\begin{document}

\title{\semihuge Multi-Modal Multi-Task Federated Foundation Models for Next-Generation Extended Reality Systems: Towards Privacy-Preserving Distributed Intelligence in AR/VR/MR}

\author[1]{Fardis Nadimi}
\author[1]{Payam Abdisarabshali}
\author[1]{Kasra Borazjani}
\author[2]{Jacob Chakareski}
\author[1*]{Seyyedali Hosseinalipour}
\affil[1]{University at Buffalo--SUNY, Department of Electrical Engineering, Buffalo, NY, USA}
\affil[2]{New Jersey Institute of Technology (NJIT), Newark, New Jersey, USA}
\affil[*]{e-mail: alipour@buffalo.edu}

\maketitle

\vspace{-4mm}

\section*{Introduction}
\begin{figure*}[ht]
    \centering
    \includegraphics[width=\textwidth]{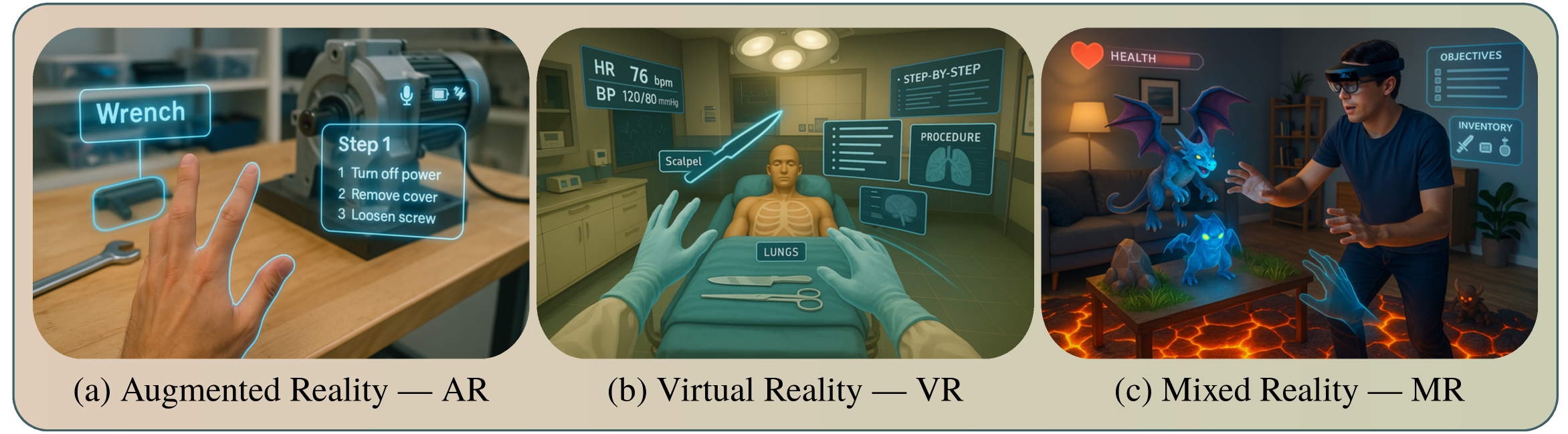}
    \vspace{-6mm}
    \caption{Various types of XR: \textbf{(a) Augmented Reality}, where digital elements are embedded into the environment surrounding a user, \textbf{(b) Virtual Reality}, where a user is fully immersed into a completely synthetic environment and interacts with various elements therein, and \textbf{(c) Mixed Reality}, which blends digital and physical worlds where a user can interact with digital objects.}
    \label{fig:xr-usecase}
\end{figure*}

Extended reality (XR) is a spectrum of immersive technologies (depicted in Fig.~\ref{fig:xr-usecase}), comprising synthetic \textit{virtual reality} (VR) scenes, \textit{augmented reality} (AR) overlays anchored to the physical world, and 
\textit{mixed/merged reality} (MR) where virtual and real objects coexist and interact with each other \cite{rauschnabel2022xr}. XR systems are envisioned as the 4\textsuperscript{th} major disruptive technology wave after PC, Internet/Web, and smartphones~\cite{TheWrap:16}, and are expected to grow their market share from \$20B in 2025 to \$123B in 2032, affecting the smartphone market share along the way \cite{FortuneBusinessInsights:23}. A defining feature of XR systems is their capacity to integrate \textit{multiple sensory modalities} in real-time, such as vision, audio, haptics, motion, and spatial mapping \cite{kourtesis2024comprehensive}. In Table~\ref{tab:xr_device_characteristics_revised_2}, we provide a structured representation of the current major XR devices and their capabilities in collecting various data modalities. This multi-modal fusion enables XR platforms to support a \textit{diverse array of downstream tasks} across various domains. In particular, XR holds tremendous potential for revolutionizing healthcare (e.g., remote surgical guidance), education (e.g., virtual lab simulations), manufacturing (e.g., real-time equipment fault detection), and entertainment (e.g., adaptive gameplay) systems \cite{ziker2021cross}. 
\begin{table*}[t]
\centering
\caption{Representative AR/VR/MR devices, along with smart glasses, and their hardware and deployment characteristics. 
\textit{Note:} Smart glasses are lightweight, often AI-powered wearables designed primarily for visual and auditory augmentation (e.g., media, notifications, translation), unlike AR/VR/MR headsets which offer fully immersive or spatially anchored environments with advanced interaction capabilities such as hand/eye tracking and spatial mapping.}
\vspace{-2mm}
\label{tab:xr_device_characteristics_revised_2}
{\notsotiny
\renewcommand{\arraystretch}{1.5}
\begin{tabular}{| > {\centering\arraybackslash}m{2.2cm} || >{\centering\arraybackslash}m{1.18cm} | >{\centering\arraybackslash}m{2.99cm} | >{\centering\arraybackslash}m{4.5cm} | >{\centering\arraybackslash}m{4.4cm}|}
\toprule
\cellcolor{darkyellow}\textbf{XR Device} & 
\cellcolor{darkyellow}\textbf{Type} & 
\cellcolor{darkyellow}\textbf{Typical Deployment Environment} & 
\cellcolor{darkyellow}\textbf{Modalities Collected} & 
\cellcolor{darkyellow}\textbf{Notes} \\
\midrule
\cellcolor{darkgreen}\textbf{Apple Vision Pro} & 
\cellcolor{darkgreen}MR & 
\cellcolor{darkgreen}Home, Offices & 
\cellcolor{darkgreen}Visual, Audio, Eye Tracking, Hand Tracking, Spatial Mapping, Motion, Depth Sensing & 
\cellcolor{darkgreen}Ultra-high resolution; advanced AR/VR passthrough \\
\midrule
\cellcolor{lightgreen}\textbf{HTC Vive XR Elite} & 
\cellcolor{lightgreen}VR / MR & 
\cellcolor{lightgreen}Home, Labs, Enterprise & 
\cellcolor{lightgreen}Visual, Audio, Motion, Hand Tracking, Depth Sensing, Spatial Mapping & 
\cellcolor{lightgreen}Modular standalone and PC VR; compact design \\
\midrule
\cellcolor{darkgreen}\textbf{Magic Leap 2} & 
\cellcolor{darkgreen}AR & 
\cellcolor{darkgreen}Industrial, Healthcare, Enterprise & 
\cellcolor{darkgreen}Visual, Audio, Eye Tracking, Hand Tracking, Spatial Mapping, Motion, Depth Sensing & 
\cellcolor{darkgreen}Lightweight AR with dynamic dimming \\
\midrule
\cellcolor{lightgreen}\textbf{Meta Quest 3} & 
\cellcolor{lightgreen}VR / MR & 
\cellcolor{lightgreen}Home, Gaming, Offices & 
\cellcolor{lightgreen}Visual, Audio, Motion, Hand Tracking, Depth Sensing, Spatial Mapping & 
\cellcolor{lightgreen}Affordable standalone with color passthrough \\
\midrule
\cellcolor{darkgreen}\textbf{Meta Quest Pro} & 
\cellcolor{darkgreen}MR & 
\cellcolor{darkgreen}Offices, Enterprise, Productivity & 
\cellcolor{darkgreen}Visual, Audio, Eye Tracking, Hand Tracking, Spatial Mapping, Motion, Depth Sensing & 
\cellcolor{darkgreen}Premium MR with enhanced passthrough and avatars \\
\midrule
\cellcolor{lightgreen}\textbf{Microsoft HoloLens 2} (discontinued in 2024)& 
\cellcolor{lightgreen}AR & 
\cellcolor{lightgreen}Hospitals, Industrial, Education & 
\cellcolor{lightgreen}Visual, Audio, Eye Tracking, Hand Tracking, Spatial Mapping, Motion, Depth Sensing & 
\cellcolor{lightgreen}Advanced AR with spatial anchors \\
\midrule
\cellcolor{darkgreen}\textbf{Ray-Ban Meta Glasses} & 
\cellcolor{darkgreen}Smart Glasses & 
\cellcolor{darkgreen}Everyday, Mobile, Social & 
\cellcolor{darkgreen}Visual, Audio & 
\cellcolor{darkgreen}Stylish glasses for immersive sensing and AI interaction \\
\midrule
\cellcolor{lightgreen}\textbf{RealWear Navigator 520} & 
\cellcolor{lightgreen}AR & 
\cellcolor{lightgreen}Industrial, Fieldwork & 
\cellcolor{lightgreen}Visual, Audio, Motion, Thermal & 
\cellcolor{lightgreen}Rugged, hands-free AR with thermal imaging option for industrial use \\
\midrule
\cellcolor{darkgreen}\textbf{Varjo XR-4} & 
\cellcolor{darkgreen}VR / MR & 
\cellcolor{darkgreen}Enterprise, Simulation, Training & 
\cellcolor{darkgreen}Visual, Audio, Eye Tracking, Hand Tracking, Spatial Mapping, Motion, Depth Sensing & 
\cellcolor{darkgreen}Human-eye resolution VR/MR; used in aviation, design, and defense training \\
\midrule
\cellcolor{lightgreen}\textbf{Vuzix Blade 2} & 
\cellcolor{lightgreen}AR & 
\cellcolor{lightgreen}Industrial, Fieldwork & 
\cellcolor{lightgreen}Visual, Audio, Motion & 
\cellcolor{lightgreen}Hands-free smart glasses with enhanced security \\
\midrule
\cellcolor{darkgreen}\textbf{XReal Air} & 
\cellcolor{darkgreen}AR & 
\cellcolor{darkgreen}Home, Travel, Entertainment & 
\cellcolor{darkgreen}Visual, Motion, Head Tracking & 
\cellcolor{darkgreen}Lightweight high-resolution AR glasses; ideal for media consumption and productivity \\
\midrule
\cellcolor{lightgreen}\textbf{Samsung Project Moohan} (upcoming) & 
\cellcolor{lightgreen}MR & 
\cellcolor{lightgreen}Home, Enterprise, Productivity & 
\cellcolor{lightgreen}Visual, Audio, Eye Tracking, Hand Tracking, Spatial Mapping, Motion, Depth Sensing & 
\cellcolor{lightgreen}Lightweight Android XR headset with Gemini AI integration and Snapdragon XR2 Gen 2\\
\midrule
\cellcolor{darkgreen}\textbf{Google AI Glasses} (upcoming) & 
\cellcolor{darkgreen}Smart Glasses & 
\cellcolor{darkgreen}Everyday, Travel, Social & 
\cellcolor{darkgreen}Visual, Audio & 
\cellcolor{darkgreen}Stylish smart glasses with in-lens display; Gemini AI for live translation and contextual assistance \\
\midrule
\cellcolor{lightgreen}\textbf{Apple Smart Glasses} (upcoming) & 
\cellcolor{lightgreen}Smart Glasses & 
\cellcolor{lightgreen}Everyday, Productivity & 
\cellcolor{lightgreen}Visual, Audio & 
\cellcolor{lightgreen}AI-powered glasses with cameras, microphones, and speakers; photo capture, translation, and navigation \\
\bottomrule
\end{tabular}
}
\end{table*}

Complementing the active research on enhancing the practicality of XR systems\cite{ChakareskiK:23,ChakareskiKRB:21,GuptaCP:20,BadnavaCH:24a}, in this work, we propose a new direction that can catalyze a major leap forward in XR intelligence: \textit{the integration of multi-modal multi-task federated foundation models}. To build a case for this integration, in the following, we articulate the motivation of this study through a curated set of guiding questions.

\begin{table*}[t]
\centering
\caption{Common ML tasks enabled by XR device data, along with associated modalities, applications, and real-time processing constraints.}
\label{tab:xr_ml_tasks}
\vspace{-1.5mm}
{\notsotiny
\renewcommand{\arraystretch}{1.5}
\begin{tabular}{| >{\centering\arraybackslash}m{2.85cm}
  || >{\centering\arraybackslash}m{2.5cm}
  | >{\centering\arraybackslash}m{4.5cm}
  | >{\centering\arraybackslash}m{4.5cm}|}
\toprule
\cellcolor{darkyellow}\textbf{ML Task} & \cellcolor{darkyellow}\textbf{Primary Modalities Used} & \cellcolor{darkyellow}\textbf{Typical Use Cases} & \cellcolor{darkyellow}\textbf{Real-Time Constraints} \\
\midrule
\cellcolor{lightgreen}\textbf{Hand Gesture Recognition} & \cellcolor{lightgreen}Visual, Motion & \cellcolor{lightgreen}Interaction control, virtual object manipulation & \cellcolor{lightgreen}Low latency (real-time interaction) \\
\midrule
\cellcolor{darkgreen}\textbf{Eye Gaze Estimation} & \cellcolor{darkgreen}Eye Tracking & \cellcolor{darkgreen}Adaptive rendering, user experience optimization & \cellcolor{darkgreen}Very low latency (frame-by-frame adjustment) \\
\midrule
\cellcolor{lightgreen}\textbf{Scene Understanding} & \cellcolor{lightgreen}Visual, Spatial Mapping & \cellcolor{lightgreen}Semantic relabeling of 3D space, navigation & \cellcolor{lightgreen}Moderate latency (background processing possible) \\
\midrule
\cellcolor{darkgreen}\textbf{Speech Recognition} & \cellcolor{darkgreen}Audio & \cellcolor{darkgreen}Voice commands, accessibility & \cellcolor{darkgreen}Low latency (near real-time) \\
\midrule
\cellcolor{lightgreen}\textbf{Activity Recognition} & \cellcolor{lightgreen}Motion, Visual & \cellcolor{lightgreen}Fitness apps, workplace monitoring & \cellcolor{lightgreen}Low to moderate latency \\
\midrule
\cellcolor{darkgreen}\textbf{Object Detection} & \cellcolor{darkgreen}Visual & \cellcolor{darkgreen}AR overlays, safety alerts & \cellcolor{darkgreen}Low latency (interactive rendering) \\
\midrule
\cellcolor{lightgreen}\textbf{User Intent Prediction} & \cellcolor{lightgreen}Eye Tracking, Motion, Audio & \cellcolor{lightgreen}Anticipatory UI, task adaptation & \cellcolor{lightgreen}Very low latency (proactive response) \\
\midrule
\cellcolor{darkgreen}\textbf{Environment Mapping} & \cellcolor{darkgreen}Visual, Spatial Mapping & \cellcolor{darkgreen}3D reconstruction, virtual overlays & \cellcolor{darkgreen}Moderate latency (updated in intervals) \\
\midrule
\cellcolor{lightgreen}\textbf{Emotion Recognition} & \cellcolor{lightgreen}Audio, Visual & \cellcolor{lightgreen}Mental health monitoring, immersive storytelling & \cellcolor{lightgreen}Moderate latency (buffered analysis acceptable) \\
\midrule
\cellcolor{darkgreen}\textbf{Augmented Object Anchoring} & \cellcolor{darkgreen}Visual, Spatial Mapping & \cellcolor{darkgreen}Persistent AR experiences, spatial memory & \cellcolor{darkgreen}Low latency (to maintain spatial stability) \\
\bottomrule
\end{tabular}
}
\vspace{-3mm}
\end{table*}
\subsubsection*{What Recent Advancements in ML Provide Unique Capabilities for Next-Generation XR Systems?}
Machine learning (ML) and subsequently multi-modal ML, techniques have become integral components of XR ecosystems\cite{wang2025towards}. To demonstrate this, in Table~\ref{tab:xr_ml_tasks}, we summarize different XR-specific ML tasks and their typically utilized data modalities.
In recent years, ML has witnessed transformative progress, owing to the emergence of \textit{foundation models (FMs)}, which are large-scale models capable of fast adaptions to a wide range of downstream tasks (e.g., via zero/few-shot learning) \cite{bommasani2021opportunities}. These models encompass a spectrum of architectures, from \textit{diffusion models} \cite{yang2023diffusion} and \textit{large language models (LLMs)} \cite{naveed2023comprehensive} to the most recent \textit{multi-modal multi-task (M3T) FMs} \cite{luo2024delving}. In particular, diffusion models (e.g., DALL·E 3) are generative models that can generate realistic data, such as images or audio. Also, LLMs such as GPT-3, BERT, and PaLM are trained on massive text corpora to predict or generate coherent language. Building on these, M3T FMs push the boundaries further by integrating \textit{multiple input modalities} (e.g., vision, language, and audio) within a unified model architecture designed to concurrently perform a \textit{variety of downstream tasks} (e.g., text/image/audio generation, and visual classification). Notable examples include PaLM-E, which combines vision and language for robotics; CLIP, which jointly learns visual and textual representations to enable various tasks; and Gato, a generalist agent capable of handling a wide range of inputs and tasks.

M3T FMs align naturally with the demands of XR environments by offering capabilities for \textit{(i) enhancing the user’s perception of their environment} and \textit{(ii) enabling the XR system's dynamic response to the user’s needs}.
Focusing on the former, M3T FMs can enrich user perception through real-time contextual overlays and semantic labeling. For instance, by processing visual and spatial inputs, M3T FMs can identify and annotate objects in the user’s field of view, translate foreign text on signage, or highlight important elements in complex environments (e.g., an antique Persian rug at a carpet exhibition). 
Focusing on the latter, by continuously interpreting multi-modal embodied signals (e.g., micro-expressions, posture, or speech patterns), M3T FMs can anticipate user needs and provide proactive support.

\subsubsection*{Is the Seamless Integration of M3T FMs into the XR Ecosystem a Trivial Endeavor?}
Despite their promising applications, integration of M3T FMs into the XR ecosystem is far from trivial and certainly not without challenges.
In particular, reaching the full potential of M3T FMs in XR often requires fine-tuning pre-trained M3T FMs (or training new ones), a process that is conventionally done at a \textit{central server} with direct access to domain/user-specific data. Nevertheless, this domain-specific data is inherently collected in a \textit{distributed} fashion at XR devices based on the real-time and continuous users’ embodied interactions (e.g., gaze, voice, motion, and even physiological). Thus, to enable FM fine-tuning or adaptation, this high-dimensional data must be continuously transferred from XR devices to centralized cloud servers; a process that places a significant burden on wireless infrastructure and raises privacy concerns associated with data leaks.

Federated learning (FL) is a recent innovation in the area of distributed ML that offers a promising way to address the above concerns \cite{mcmahan2017communication}. FL enables collaborative model training directly on user devices, eliminating the need to share their raw data with central servers. 
Integrating M3T FMs within the FL paradigm gives rise to a novel framework, called \textit{M3T Federated Foundation Models (FedFMs)}, which enables collaboratively trained M3T FMs across various XR devices while alleviating the privacy concerns \cite{qiao2025towards, chen2024feddat}.
The decentralized training of M3T FedFMs across XR devices further unlocks the potential for \textit{personalizing local M3T FMs to individual user and device characteristics}.
Nevertheless, integrating M3T FedFMs into XR ecosystems would encounter multi-dimensional challenges, the identification of which is one of the goals of this paper.

\subsubsection*{What Signals Point to the Feasibility of Integrating M3T FedFMs into the XR Ecosystem?}
We note that our vision for integrating M3T FedFMs into the XR ecosystem is motivated by the emerging body of work that demonstrates the successful implementation of their components (i.e., FL and M3T FedFMs). In particular, on one hand FL has gained traction in a variety of domains, such as mobile health \cite{wang2023applications}, Internet-of-Things (IoT) \cite{savazzi2020federated}, Industrial IoT (IIoT) \cite{nguyen2021federated}, vehicular networks \cite{elbir2022federated}, and in non-terrestrial networks \cite{han2024cooperative}.
In XR settings, FL is particularly appealing: XR devices generate continuous streams of sensitive and personalized data (e.g., gaze trajectories, hand gestures, motion patterns), where FL can enable on-device model training while preserving user privacy. 
On the other hand, FMs and M3T FMs have emerged as transformative technologies both in XR and other domains \cite{moor2023foundation,afzal2025next,li2022novel}. Specifically, in XR, at the system level, early efforts from Meta Reality Labs \cite{MetaRealityLabs} and NVIDIA Omniverse \cite {NVIDIAOmniverse} demonstrate XR-focused M3T FM pipelines and simulation platforms. Yet all of these implementations still rely on conventional \textit{centralized} M3T FMs. To further provide a better understanding of the uniqueness of XR settings, Table~\ref{tab:fl_fm_xr_limitations} provides an overview of the common limitations of the existing studies on M3T FM and FL when extended to M3T FedFM model training for XR applications.

Although training challenges and stringent inference requirements, especially in XR environments, pose potential hurdles for FedFMs, recent trends suggest a promising trajectory. FedFMs and M3T FedFMs are becoming increasingly practical, as evident in other fields \cite{borazjani2025multi, du2024distributed, li2025open}, driven by recent advances in resource-efficient local FM training/adaptation techniques and the emergence of compact, high-efficiency models such as Google DeepMind’s Gemma 3n \cite{google}, Meta Llama 3.2 \cite{meta}, and Apple Intelligence \cite{apple}, which are specifically engineered for real-time, on-device multi-modal inference. Further, in the AI/ML community, there is high interest in advancing parameter-efficient training/fine-tuning techniques, such as low rank adaptation (LoRA) and adapter modules, making the operations of M3T FedFMs over resource-constrained wireless devices increasingly practical \cite{zhuang2023foundation}.

\begin{tcolorbox}[colback=lightgreen,colframe=darkgreen!75!black,colbacktitle=red!80!black,
title=\centering \small The Vision and Goal of this Study,fonttitle=\bfseries]
Despite their tremendous potentials and compelling motivations, the integration of M3T FedFMs within the XR ecosystem remains unexplored, primarily due to the emerging nature of M3T FedFMs themselves. This gap forms the central motivation for this study, in which we \textbf{(1)} identify application scenarios for M3T FedFMs in XR, \textbf{(2)} articulate the specific challenges facing the deployment of M3T FedFMs in XR environments, and \textbf{(3)} propose a performance evaluation framework for M3T FedFMs tailored to XR-specific needs, focusing on both the XR applications performance and the resource-efficiency of FedFM training/deployment. Collectively, our aim is to catalyze further research in this untapped space and lay the groundwork for realizing private, adaptive, and scalable intelligence within XR systems.

\end{tcolorbox}

\begin{table*}[t]
\centering
\caption{Common limitations of FL and M3T FM approaches in XR contexts. Note that exceptions exist in each of the presented limitations, and the table rather presents the ``common theme" of limitations.}
\label{tab:fl_fm_xr_limitations}
\vspace{-1.5mm}
{\notsotiny
\renewcommand{\arraystretch}{1.5}
\begin{tabular}{| > {\centering\arraybackslash}m{2cm}
  || >{\centering\arraybackslash}m{4.1cm}X
| >{\centering\arraybackslash}m{4cm}X
  | >{\centering\arraybackslash}m{5.5cm}|}
\toprule
\cellcolor{darkyellow}\textbf{Dimension} & \cellcolor{darkyellow}\textbf{Traditional FL} & \cellcolor{darkyellow}\textbf{Conventional M3T FMs} & \cellcolor{darkyellow}\textbf{XR-Specific Challenges} \\
\midrule
\cellcolor{lightgreen}\textbf{Modality Diversity} & \cellcolor{lightgreen}Assumes uniform data modalities across clients; struggles with clients having different or missing modalities. & \cellcolor{lightgreen}Pretrained on fixed modality combinations; lacks seamless adaptation to new or missing modalities without centralized fine-tuning. & \cellcolor{lightgreen}XR devices capture diverse and dynamic modalities (e.g., vision, audio, haptics), which require handling modality heterogeneity and intermittent modality dropout during sessions. \\
\midrule
\cellcolor{darkgreen}\textbf{Task Variability} & \cellcolor{darkgreen}Designed for homogeneous tasks; performance degrades with task heterogeneity among clients. & F\cellcolor{darkgreen}ine-tuned for specific tasks; adapting to new tasks requires significant computational resources and data collection. & \cellcolor{darkgreen}XR applications involve diverse and rapidly changing tasks (e.g., navigation, interaction, recognition), which demand adaptation to task variability across users and over time. \\
\midrule
\cellcolor{lightgreen}\textbf{Real-Time Adaptation} & \cellcolor{lightgreen}Limited support for real-time model updates; communication overhead hinders low-latency requirements. & \cellcolor{lightgreen}Large model sizes and inference times impede real-time responsiveness; not optimized for low-latency environments. & \cellcolor{lightgreen}XR systems demand immediate adaptation to user interactions and environmental changes, which necessitate low-latency model updates. \\
\midrule
\cellcolor{darkgreen}\textbf{Hardware Constraints} & \cellcolor{darkgreen}
Often trains conventional ML models, where computation constraints may not be as restricting compared to FMs. & \cellcolor{darkgreen}Require substantial memory, processing power, and energy supply; unsuitable for deployment on resource-limited devices. &\cellcolor{darkgreen} XR devices often have limited computational capabilities, which makes it challenging to deploy and train/deploy large models efficiently. \\
\midrule
\cellcolor{lightgreen}\textbf{Privacy Preservation} & \cellcolor{lightgreen}Provides data privacy by design; however, model updates can still leak sensitive information unless security countermeasures (e.g., differential privacy or encryption) are used. & \cellcolor{lightgreen}Centralized training poses privacy risks. & \cellcolor{lightgreen}XR applications collect sensitive user data (e.g., location), which requires robust privacy-preserving mechanisms during model training and inference. \\
\midrule
\cellcolor{darkgreen}\textbf{Personalization} & \cellcolor{darkgreen}Personalization is limited and often requires additional mechanisms; struggling with balancing global and local model performance. & \cellcolor{darkgreen}Fine-tuning for personalization is often not feasible due to not having access to the users' private data; may not generalize well across users. &\cellcolor{darkgreen} XR experiences are highly personalized, which calls for local models that can adapt to individual user preferences and behaviors in real-time. \\
\midrule
\cellcolor{lightgreen} \textbf{Model Aggregation} & \cellcolor{lightgreen}Standard aggregation methods assume homogeneous models and data distributions; ineffective with heterogeneous client models and tasks. &\cellcolor{lightgreen} Aggregation across different FMs is complex due to varying architectures and training objectives. & \cellcolor{lightgreen}In XR, clients may have diverse models and tasks, which requires advanced aggregation techniques that can handle heterogeneity in both model structures and user objectives. \\
\midrule
\cellcolor{darkgreen}\textbf{Communication Overhead} & \cellcolor{darkgreen}High communication costs due to frequent model updates. & \cellcolor{darkgreen}Models are trained locally with no aggregations. & \cellcolor{darkgreen}XR devices often generate high volumes of data and operate under battery constraints and limited bandwidth, which necessitates communication-efficient model update strategies. \\
\midrule
\cellcolor{lightgreen}\textbf{Security and Robustness} & \cellcolor{lightgreen}Vulnerable to adversarial attacks and model poisoning; lacks robust mechanisms to ensure model integrity. & \cellcolor{lightgreen} Often trained and kept in a central server equipped with firewalls & \cellcolor{lightgreen} Distributed XR systems comprise unpredictable environments and users, which require models that are robust to adversarial conditions and can maintain integrity over time. \\
\bottomrule
\end{tabular}
}
\end{table*}

\section*{M3T FedFMs: Model Architecture, Network Orchestration, and Training Strategy
}

In its conventional form, FL conducts collaborative model training across devices by repeating two iterative operations until model convergence: (i)  devices train their local models on their local datasets, and (ii) devices engage in transmitting their trained models (either to a server or among themselves) to aggregate their models and will use the aggregated model to continue their local training.
Likewise, M3T FedFMs entail the orchestration of the training/fine-tuning of \textit{local M3T FMs at the devices} through FL operations. 
Subsequently, understanding M3T FedFMs requires knowledge of M3T FMs and FL working mechanisms, which we provide next.

\begin{figure*}[t]
    \centering
    \includegraphics[width=\textwidth]{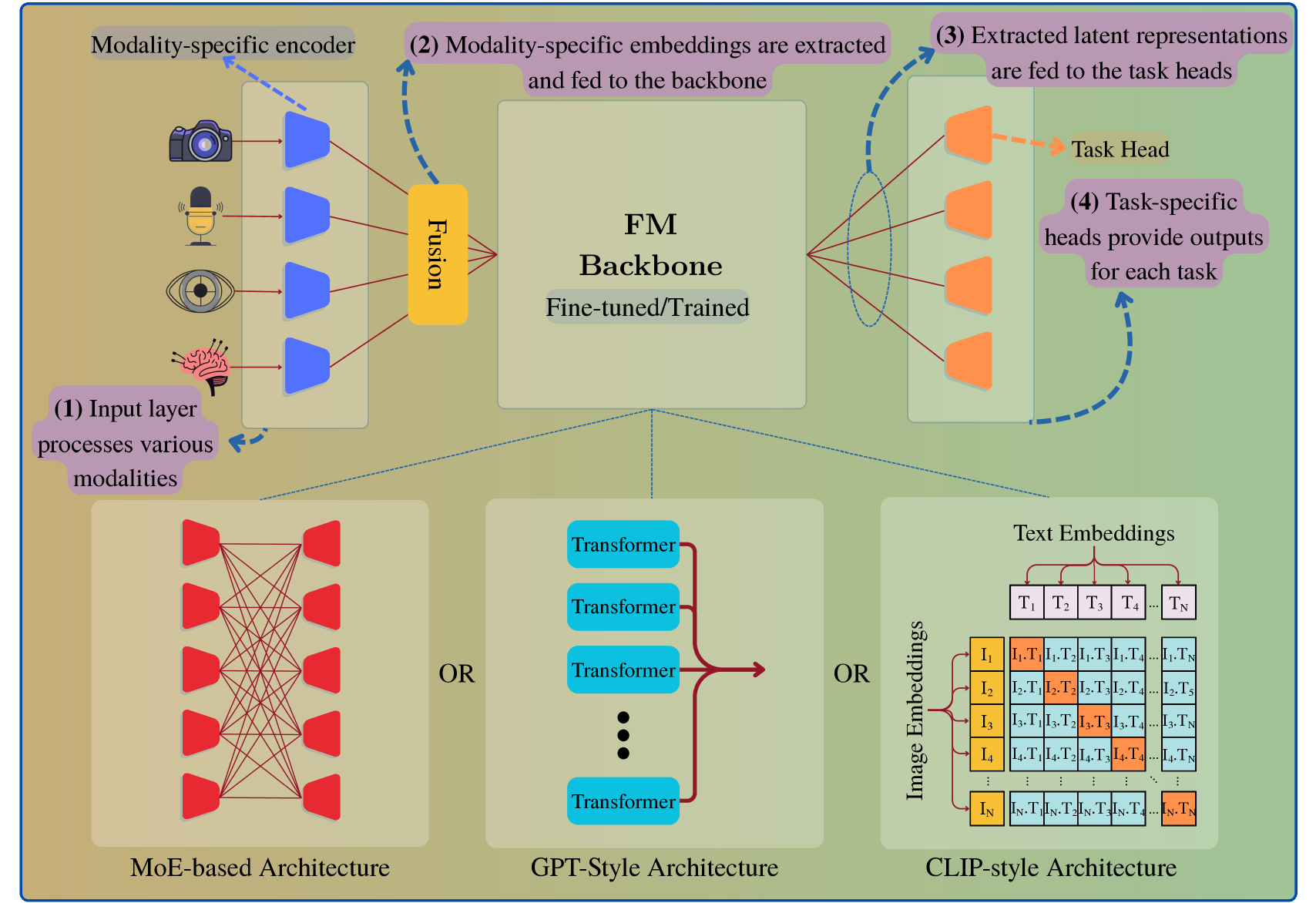}
    \vspace{-7mm}
    \caption{Foundation model architecture containing the input layer (i.e., modality encoders), the FM backbone, and the output layer (i.e., task heads). The FM backbone can comprise varying architectures, such as Mixture-of-Experts (MoE), GPT-style (stacked transformers), and CLIP-style (contrastive learning-based, where transformers are used for image and text processing) architectures.} 
    \label{fig:foundational_models}
\end{figure*}

We note that M3T FMs are not defined by a single unified architecture but rather encompass a family of designs under active exploration in both academia and industry. In the following, we abstract a set of popular M3T FM architectures (depicted in 
Fig.~\ref{fig:foundational_models}) that are inspired by current trends in the research area \cite{chen2024disentanglement, xiao2024configurable}, followed by the model training procedure of FL.

\noindent\textbf{1. Modality Encoders:} 
M3T FMs use dedicated \textit{modality encoders} that convert raw sensory inputs into standardized intermediate embeddings.  
These encoders operate \textit{prior to fusion}, serving as the entry point for the model backbone.

\noindent\textbf{2. Shared Backbone:} 
After the initial sensory encoding, the intermediate modality-specific embeddings are passed to a shared backbone, which performs multi-modal fusion, reasoning, and contextual learning. This backbone is central to generalization across tasks and users, and it can take several architectural forms depending on the model's design goal. These architectural forms are depicted at the bottom of Fig.~\ref{fig:foundational_models} and detailed below:

\begin{itemize}[leftmargin=3mm, itemsep=-0.25em]
    \item \textit{Transformer Backbones (e.g., GPT-style) \cite{brown2020language}:} A stack of transformers process all input embeddings. 

    \item \textit{Dual-Encoder or Multi-Encoder Fusion (e.g., CLIP-style) \cite{radford2021learning}:} Each modality is processed by its own encoder (or backbone stream, which can be a transformer), and the outputs are aligned or fused in a shared latent space, often via contrastive learning or cross-attention. 

\item \textit{Mixture-of-Experts (MoEs) \cite{chen2024disentanglement}:} 
MoEs refer to a collection of parallel expert sub-networks, often transformer blocks or neural networks, selectively activated depending on the input modality or the targeted tasks. MoEs play two roles: (1) selecting the most relevant set of modality combinations for processing, and (2) routing the resulting fused representations through task-relevant expert pathways before reaching the output heads. Inside the backbone, a gating network determines which subset of experts to activate per input instance, allowing the model to maintain computational efficiency. 
\end{itemize}

\begin{table*}[t]
\centering
\caption{Pairwise comparison of M3T FM local training strategies. Each cell explains how the row method differs from the column method in terms of model behavior modification, modality/task adaptability, and hardware feasibility.}
\label{tab:local_training_comparison_detailed}
\vspace{-1.5mm}
{\notsotiny
\renewcommand{\arraystretch}{2.0}
\begin{tabular}{| >{\centering\arraybackslash}m{1.55cm}
|| >{\centering\arraybackslash}m{2.7cm}
| >{\centering\arraybackslash}m{2.6cm}
| >{\centering\arraybackslash}m{2.2cm}
| >{\centering\arraybackslash}m{3.2cm}
| >{\centering\arraybackslash}m{2.5cm}|}
\toprule
\cellcolor{darkyellow}\textbf{Compared To $\rightarrow$} & 
\cellcolor{darkyellow}\textbf{Prompt Tuning} & 
\cellcolor{darkyellow}\textbf{Adapter Tuning} & 
\cellcolor{darkyellow}\textbf{MoE Expert Tuning} & 
\cellcolor{darkyellow}\textbf{Full Transformer / Backbone Tuning} & 
\cellcolor{darkyellow}\textbf{LoRA Tuning} \\
\midrule

\cellcolor{lightgreen}\textbf{Prompt Tuning} & 
\cellcolor{lightgreen}--- & 
\cellcolor{lightgreen}Consumes fewer parameters and memory, but cannot significantly adapt to new modalities or tasks. & 
\cellcolor{lightgreen}Lightweight but applies a fixed prompt across all inputs, limiting adaptation to diverse modalities. & 
\cellcolor{lightgreen}Can run on resource-constrained devices, but lacks depth and task adaptation capacity provided by full fine-tuning. & 
\cellcolor{lightgreen}Less expressive than LoRA, which introduces learnable rank-specific updates that better balance adaptability and footprint. \\
\midrule

\cellcolor{darkgreen}\textbf{Adapter Tuning} & 
\cellcolor{darkgreen}Can better tailor responses to new tasks or sensors than prompt tuning, but needs slightly more compute/memory. & 
\cellcolor{darkgreen}--- & 
\cellcolor{darkgreen}Applies uniformly across input, whereas MoE can adapt per sample with selective expert routing. & 
\cellcolor{darkgreen}Offers reasonable task coverage with less resource overhead compared to tuning all transformer weights. & 
\cellcolor{darkgreen}LoRA is lighter than adapters in compute and memory (and inference) but slightly less modular for some downstream tasks. \\
\midrule

\cellcolor{lightgreen}\textbf{MoE Expert Tuning} & 
\cellcolor{lightgreen}Dynamically activates specialized sub-modules per input type, boosting adaptability at the cost of routing complexity and memory. & 
\cellcolor{lightgreen}More modular and task-adaptive in theory, but needs well-designed expert architecture to be efficient. & 
\cellcolor{lightgreen}--- & 
\cellcolor{lightgreen}Often achieves similar performance to full model tuning if expert selection is well optimized, but with significantly lower resource costs. & 
\cellcolor{lightgreen}LoRA is more memory-efficient but lacks the task-specific routing of MoE; MoE suits structured modality switching better. \\
\midrule

\cellcolor{darkgreen}\textbf{Full Transformer / Backbone Tuning} & 
\cellcolor{darkgreen}Deeply retrains all parameters for strong adaptation, but impractical for edge device constraints. & 
\cellcolor{darkgreen}Too resource-heavy for real-time or on-device use; adapter tuning is more suitable for low inference-time adaptation. & 
\cellcolor{darkgreen}Consumes more memory/compute than MoE without necessarily improving adaptability in structured tasks. & 
\cellcolor{darkgreen}--- & 
\cellcolor{darkgreen}LoRA achieves comparable performance in many cases with drastically fewer trainable parameters and better deployability. \\
\midrule

\cellcolor{lightgreen}\textbf{LoRA Tuning} & 
\cellcolor{lightgreen}Provides rank-based updates that adapt better than fixed prompts while retaining low overhead. & 
\cellcolor{lightgreen}Often more efficient in training/inference compared to adapters, but slightly less flexible for multi-modal fusion. & 
\cellcolor{lightgreen}Lacks dynamic expert routing but excels in memory- and compute-limited settings. & 
\cellcolor{lightgreen}Can mimic some full-tuning benefits with reduced cost, making it viable for on-device learning. & 
\cellcolor{lightgreen}--- \\
\bottomrule
\end{tabular}
}
\end{table*}

\noindent\textbf{3. Task Heads:} 
Task heads in the M3T FM architecture aim to map the shared representations, outputted from the backbone, to task-specific outputs.

\noindent\textbf{4. Fine-Tuning Capabilities (Adapters and Prompts):} 
To support efficient personalization in resource-constrained devices, M3T FMs often integrate lightweight adaptation mechanisms, namely, \textit{adapters} and \textit{prompt-based tuning}, defined below.

\begin{itemize}[leftmargin=3mm, itemsep=-0.25em]
    \item \textit{Adapters} refer to small, trainable modules inserted between or within layers of the shared backbone. These modules can be trained at each device to fine-tune the entire FM rather than updating all of its parameters. Adapters enable each device to fine-tune the model's behavior locally while keeping the majority of the backbone frozen, thereby reducing computation and energy burden \cite{long2024dual}.
    \item \textit{Prompts} are learnable input tokens, obtained via \textit{prompt-tuning}, prepended to the sequence of modality embeddings or inserted at key positions within the network. Rather than modifying internal weights, prompts steer model behavior to personalize the local model to the device, making them ideal for low-latency on-device updates \cite{jia2022visual}.
\end{itemize}

\noindent\textbf{5. FL Coordination Paradigm:} Following FL principle, in M3T FedFM, each device performs local training over its dataset to update its local M3T FM modules/parameters; however, \textit{this process usually does not entail the training of the entire M3T FM}, as the burden of such training exceeds the capabilities of resource-constrained devices. Instead, the local training is often performed in the form of prompt-tuning, learning the adapter parameters (e.g., via LoRA-based methods \cite{ghiasvand2025few}), or
updating relevant M3T FM components, such as local experts. In Table~\ref{tab:local_training_comparison_detailed}, we provide a pairwise comparison between the techniques that can be used during the local training, identifying their unique advantages and disadvantages. 

Followed by local training, the devices can use three various FL coordination paradigms (depicted in Fig.~\ref{fig:federated-learning}), which are detailed below and compared against each other in Table~\ref{tab:ffm_training_aggregation}, to improve their local model generalizability and performance:
\begin{figure*}[!h]
    \centering
    \includegraphics[width=\textwidth]{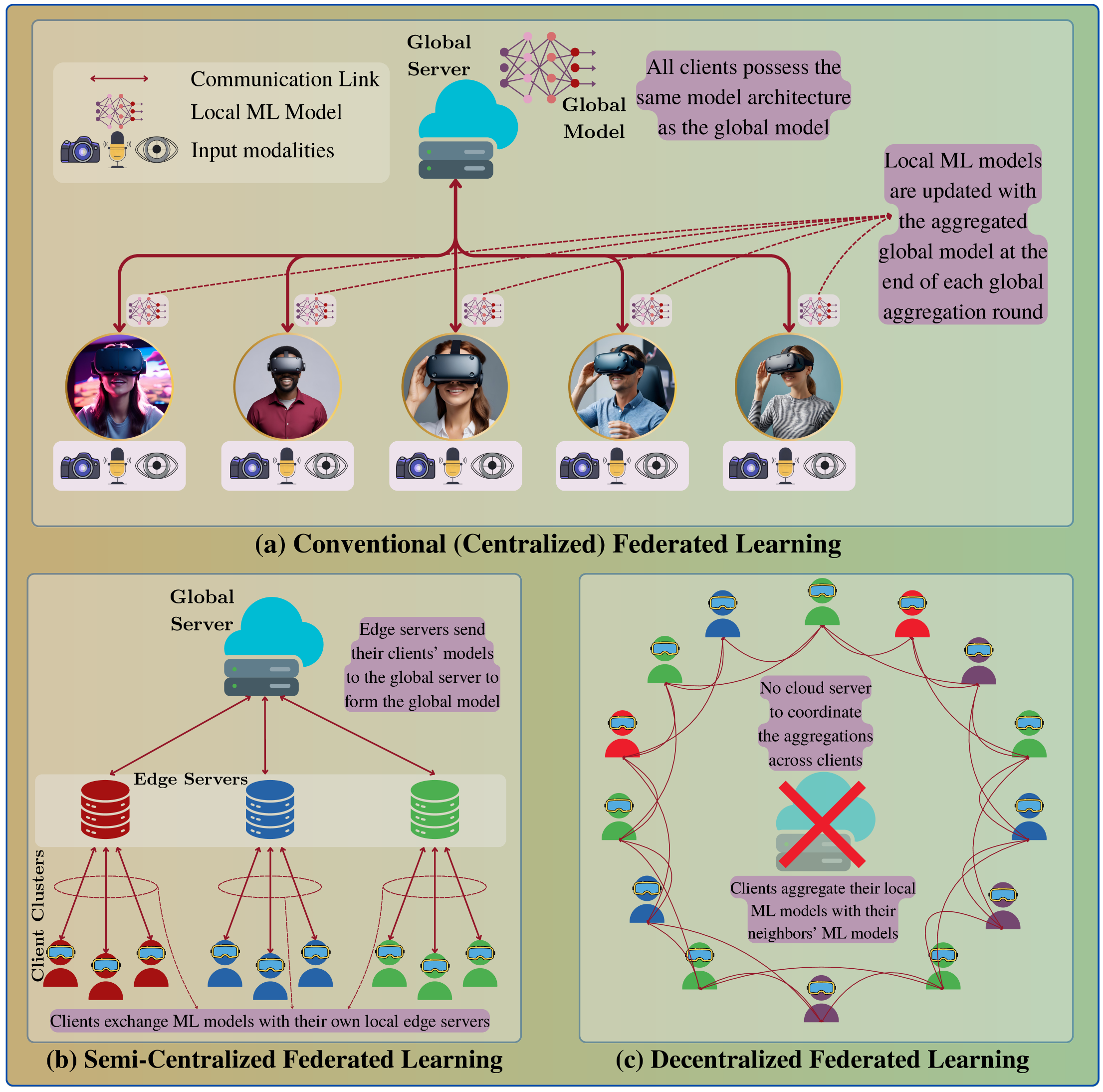}
    \vspace{-7mm}
    \caption{Various model aggregation and network coordination strategies in FL: \textbf{(a) Conventional FL} employs a centralized model aggregation strategy where clients perform local training and send their model updates to a central server. The server aggregates these updates into a global model and broadcasts it back to the clients.
\textbf{(b) Semi-decentralized FL} introduces a hierarchical structure where clients are organized into clusters. Within each cluster, clients exchange models via D2D communications and reach local consensus \cite{parasnis2024energy,abdisarabshali2024dynamic}. A selected client from each cluster then sends the aggregated model to the central server, which constructs a global model and distributes it across all clients.
\textbf{(c) Fully decentralized FL} operates without a central server; model updates are exchanged entirely over D2D links, where each client updates its local model based on models received from its neighbors (e.g., using distributed consensus algorithms).}
    \label{fig:federated-learning}
    \vspace{-3mm}
\end{figure*}
\begin{itemize}[leftmargin=3mm, itemsep=-0.25em]
    \item \textbf{Centralized Aggregation:} A cloud server collects model updates from all participating devices. It then aggregates these updates to form a global model, typically via weighted averaging, and redistributes/broadcasts the global model back to the devices. This style of aggregation, although it offers good scalability, engages devices in energy-intensive uplink communications, making it less ideal for implementation over resource-limited systems \cite{mcmahan2017communication}.

    \item \textbf{Fully Decentralized Aggregation:} In the absence of any central server, devices collaborate purely through energy-efficient, short-range device-to-device (D2D) model exchanges, the aggregation is often done through distributed consensus mechanisms.
    While this aggregation method provides maximum autonomy and energy-efficiency, it introduces challenges in convergence and is often not suitable for large-scale networks as it requires a connected network substrate formed by the D2D links \cite{zehtabi2024decentralized}.

    \item \textbf{Semi-Decentralized Aggregation:} Devices are organized into clusters based on spatial proximity, connectivity, or task similarity. Within each cluster, devices engage in energy-efficient D2D communications to reach a local model consensus. Afterwards, one device, often selected dynamically, acts as a cluster representative and performs an uplink to the central server. The server then forms the global model through the received updates, as in the centralized aggregation, and broadcasts the global model back to the devices. This approach reduces uplink communication overhead and can promote local personalization of models at the cluster level; however, it is only deployable when devices are capable of D2D communications \cite{lin2021semi}.
\end{itemize} 

\noindent\textbf{6. Training Strategies of M3T FedFMs:} Effective training of M3T FedFMs demands a fundamental reassessment of conventional FL techniques, given the architectural and functional shifts introduced by M3T models. Techniques such as model distillation, quantization, sparsification, intelligent client selection, and parameter-efficient fine-tuning (PEFT), such as LoRA, have been predominantly developed for monolithic, task-specific deep neural networks or LLMs, where their extensions to the modular and multi-modal multi-task nature of emerging M3T FedFMs remain highly underexplored \cite{kuo2024federated, long2024dual}. A promising direction lies in adapting and extending these techniques to align with the modular structure of M3T models. Specifically, their modular design, featuring expert modules, task-specific heads, and lightweight adapters, enables selective/module-wise training and fine-tuning on client devices, followed by modular-level aggregations, which calls for the revisiting of these existing techniques. For instance, model distillation (traditionally used to compress a large model into a smaller one) can evolve into a modality-aware distillation process that preserves knowledge across different sensor modalities and tasks without collapsing their distinct representations. Similarly, quantization and sparsification can be applied selectively to different modules (e.g., encoders vs. task heads), balancing compression with model fidelity. Also, intelligent client selection (typically relying on system metrics such as availability or data quality) can be expanded to incorporate modality-task coverage, prioritizing clients that possess underrepresented sensor combinations or task labels. Finally, PEFT methods such as LoRA or adapters (originally designed for sequential fine-tuning of LLMs) should be reconfigured to support multi-branch, modular updates (i.e., updates targeting specific components of the model such as modality-specific encoders, task heads, or expert modules), allowing clients to locally fine-tune only relevant modules and transmit them for federated aggregation. Taking this one step further, when deploying M3T FedFMs in XR systems, these revamped techniques must also account for the unique interaction, resource, and environment characteristics of XR, which we will later elaborate on through the \textbf{SHIFT} dimensions in the following sections.

\begin{table*}[t]
\centering
\caption{Comparison of data/model aggregation mechanisms for M3T FedFMs in XR ecosystems.}
\label{tab:ffm_training_aggregation}
\renewcommand{\arraystretch}{1.5}
\vspace{-1.5mm}
{\notsotiny
\begin{tabular}{| >{\centering\arraybackslash}m{2.8cm} 
|| >{\centering\arraybackslash}m{3.8cm} 
| >{\centering\arraybackslash}m{4.5cm} 
| >{\centering\arraybackslash}m{4.5cm}|}
\toprule
\cellcolor{darkyellow}\textbf{Strategy} & \cellcolor{darkyellow}\textbf{Definition} & \cellcolor{darkyellow}\textbf{Benefits} & \cellcolor{darkyellow}\textbf{Drawbacks} \\
\midrule

\cellcolor{lightgreen} \textbf{Centralized Training} & \cellcolor{lightgreen}
All training data is collected and processed on a central server. & \cellcolor{lightgreen}
Simplifies coordination; supports large-scale pretraining; ensures global consistency. & \cellcolor{lightgreen}
Raises privacy concerns; assumes constant connectivity; not ideal for on-device personalization. \\
\midrule

\cellcolor{darkgreen} \textbf{Centralized FL} & \cellcolor{darkgreen}
Devices train locally and send model updates to a central server for aggregation. & \cellcolor{darkgreen}
Supports privacy; adapts to local data distributions; scalable. & \cellcolor{darkgreen}
Relies on server availability; Stragglers (e.g., low-compute users) can affect the model convergence. \\
\midrule

\cellcolor{darkgreen}\textbf{Fully Decentralized FL} & \cellcolor{darkgreen}
Devices average their models via D2D consensus without relying on any central server. & \cellcolor{darkgreen}
Resilient against server failures/downtimes; useful in ad-hoc or low-connectivity networks. & \cellcolor{darkgreen}
Slow convergence; harder to maintain model consistency across devices. \\
\midrule

\cellcolor{lightgreen}\textbf{Semi-Decentralized FL} & \cellcolor{lightgreen}
Devices form clusters to reach local D2D consensus on their models before one representative communicates with the server. & \cellcolor{lightgreen}
Reduces uplink bandwidth; supports localized adaptation; balances load across the network. & 
\cellcolor{lightgreen} Adds coordination complexity within clusters. \\
\bottomrule
\end{tabular}
}
\end{table*}

\section*{M3T FedFMs Use Cases in the XR Ecosystem}
\begin{figure*}[t]
    \centering
    \includegraphics[width=\textwidth]{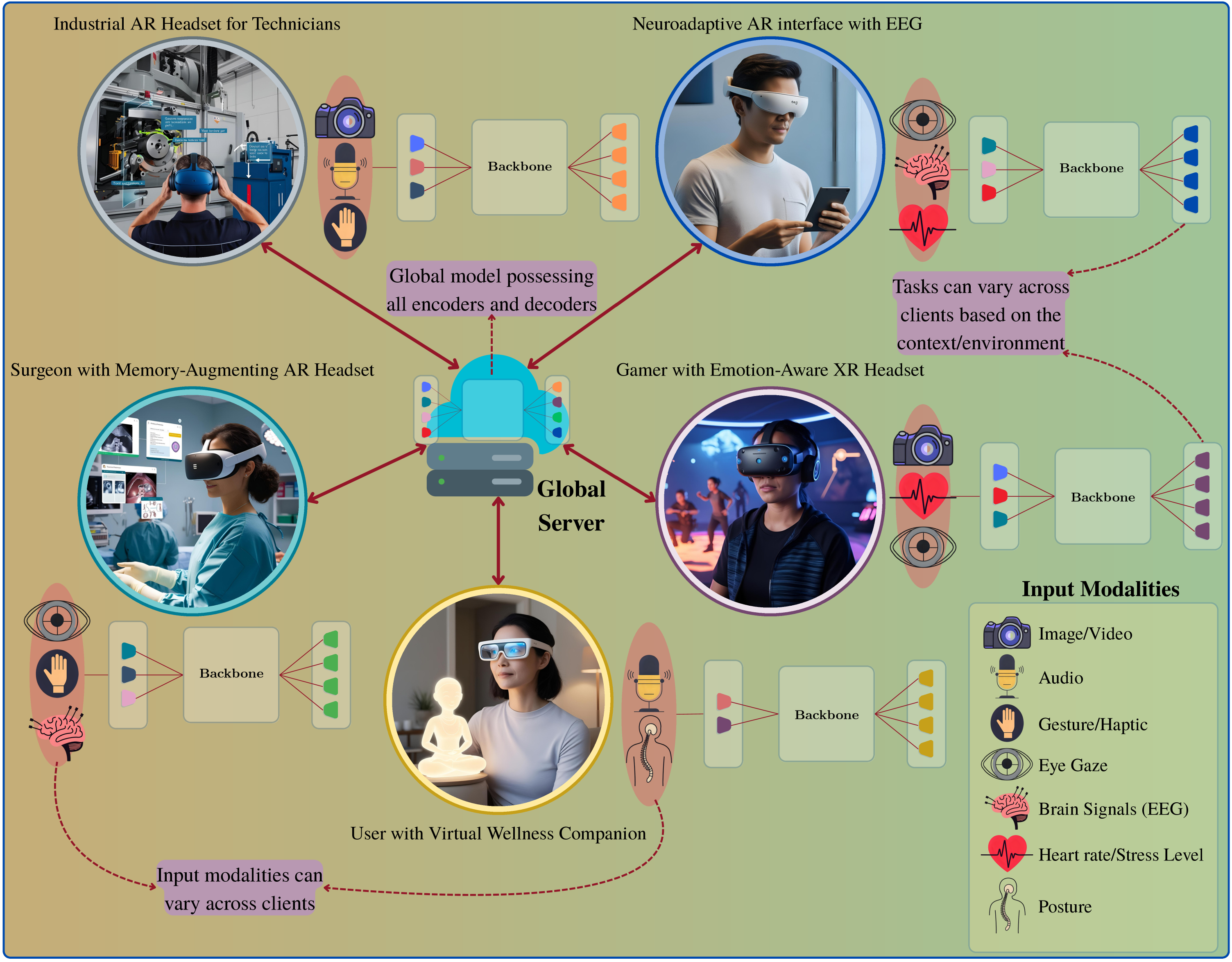}
    \vspace{-7mm}
    \caption{FedFM deployment visualization in an XR system, under FL with centralized model aggregation strategy. Different deployment environments of XR devices will result in varying tasks and modalities across the clients.}
    \label{fig:ffm}

    \vspace{-5mm}
\end{figure*}

In the following, we highlight a range of scenarios where M3T FMs have the potential to redefine how XR systems perceive, adapt, and interact, spanning domains such as healthcare, education, manufacturing, and entertainment. These scenarios, schematically visualized in Fig.~\ref{fig:ffm}, represent a blend of evolved versions of existing XR deployments and forward-looking use cases that push the boundaries of what is possible. Notably, with the recent wave of on-device FM integration into smartphones \blue{~\cite{google, meta, apple}}, it is increasingly clear that no major player in the XR space, including tech giants such as Apple, Google, and Meta can afford to ignore the paradigm shift that the recently emerged M3T FMs, and by extension, M3T FedFMs, are poised to bring to this field.

We emphasize that in presenting the scenarios, our focus is on justifying the \textit{use of well-trained/fine-tuned} M3T FMs within each context. The discussion on \textit{how these models are obtained in the first place} follows later in the section, anchored by a key observation: M3T FMs intended for deployment on XR devices cannot be feasibly trained via conventional centralized pipelines due to constraints in privacy, bandwidth, and personalization. Instead, these models must be trained through FL approaches, highlighting the essential role of M3T FedFMs across all the scenarios presented.

\subsubsection*{(Scenario 1) Cognitive XR Augmentation: Real-Time Perceptual Overlays}
Cognitive XR augmentation entails using the multi-modal embodied data to generate useful, real-time, perceptual overlays for various downstream tasks \cite{rajan2024heads}.
In these settings, the multi-modal embodied data includes visual (e.g., scene context and handwriting), audio (e.g., spoken commands), and user-specific cues (e.g., emotional tone). Also, the typical downstream tasks are live translation, contextual labeling of objects, and environment-aware recommendations.

In such environments, M3T FMs are particularly potent due to their ability to handle various modalities and adapt to multiple concurrent tasks. For instance, a user might look at a handwritten draft and instantly receive suggestions on clarity, grammar, or tone, all rendered directly within their visual field through a well-trained M3T FM \cite{xu2019design}. Similarly, in industrial settings, a technician wearing an XR headset may receive dynamic overlays pointing out abnormal machine vibrations, heat signatures, or wear patterns, all interpreted and delivered by an M3T FM that feeds the overlays to the headset \cite{raj2024augmented}.

\subsubsection*{(Scenario 2) Neuroadaptive Interfaces: XR with Brain-Computer Integration}
Neuroadaptive XR interfaces represent the next frontier in embodied interaction, where XR systems respond to both the external behavior of users and their internal cognitive and emotional states.
These interfaces have come closer to reality owing to the advances in neural sensing technologies, including wearable electroencephalogram (EEG) devices, functional near-infrared spectroscopy (fNIRS) sensors, and implantable brain–computer interfaces (BCIs) such as Neuralink \cite{musk2019integrated}, which collectively make the integration of brain activity as a live input stream within XR environments a feasibility. 

In such settings, M3T FMs can enable truly \textit{intention-driven experiences}, an emerging paradigm toward which both industry and academia are increasingly converging \cite{zhu2024human}. For instance, a user might simply \textit{think} about a command on opening an application, and the interface seamlessly responds without requiring any spoken or gestural command. In the medical domain, this capability holds transformative potential \cite{vourvopoulos2019effects, nag2024tumorganet}. 
For example, a patient with motor impairments might imagine moving a paralyzed limb. Subsequently, the M3T FM, trained to process multi-modal embodied inputs (e.g., neural signals, visual context, and spatial mapping), can interpret this mental intent and translate it into the movement of a virtual limb within the XR environment \cite{vourvopoulos2019efficacy}.

\subsubsection*{(Scenario 3) Embodied AI Companions: Adaptive Agents}
Another future application of XR systems, actively pursued in both academia and industry, is the creation of embodied AI companions \cite{yousri2024illusionx}. These AI companions are intelligent agents that evolve alongside the user. These agents can manifest as visual avatars or operate invisibly in the background, providing contextual support such as reminders, co-navigation, or behavioral nudges \cite{lima2025real}. The operation of these systems is grounded in the continuous acquisition of multi-modal embodied data streams, including prosodic/rhythmic features of speech (e.g., pitch, cadence) and ocular patterns. Also, the computational tasks associated with such systems encompass emotional state classification, fine-grained intent recognition, longitudinal behavior modeling, and real-time context-aware recommendation generation \cite{merrill2022ai}. 

In these systems, M3T FMs can lead to unprecedented advancements. For example, a companion powered by an M3T FM may detect a gradual decline in the user's emotions, moods, or feelings  (e.g., based on combined analysis of speech, gaze, and posture), and preemptively offer mental health resources \cite{eerdenisuyila2024analysis, jafaridesign}. Also, by leveraging contextual cues such as eye-gaze and user behavior, AI companions can provide suggestions in terms of overlays to improve the user's productivity \cite{bovo2025embardiment}.

\subsubsection*{(Scenario 4) Emotionally Resonant Storytelling in XR Gaming}
With modern XR headsets equipped with sensors for gaze tracking, pupil dilation, facial micro-expressions, and vocal stress detection, \textit{emotion-aware gaming} can transition from a futuristic concept to an emerging reality, the promise of which has triggered a major interest in the gaming industry \cite{lopes2025closing}. XR devices can continuously collect multi-modal embodied signals that reflect the player's internal/emotional state in real-time \cite{kiafar2025mena}. Building on this capability, adaptive storytelling is emerging as a new frontier in emotion-aware gaming, where narratives evolve not only in response to a player's actions, but also in response to their physiological states \cite{savacs2024biofeedback, klein2023exploring}. 
This enhances the emotional depth that XR games can offer, fostering a more profound and personalized connection between players and virtual environments. Notably, significant efforts have been directed toward dynamically eliciting/triggering complex emotions such as fear, creating deeply immersive and psychologically resonant gaming experiences \cite{de2020towards,zaib2022using}.

In this context, M3T FMs' ability to integrate diverse sensory inputs and simultaneously handle multiple downstream tasks makes them ideal for dynamically adjusting the pace, intensity, and structure of game narratives. Specifically, a well-trained M3T FM can recognize when a player is deeply immersed, emotionally overwhelmed, or disengaged, and adapt the storytelling rhythm accordingly.

\begin{tcolorbox}[colback=lightgreen,colframe=darkgreen!75!black,colbacktitle=red!80!black,
title=\centering \small Why in the Above (and All Comparable) Scenarios M3T FMs Must be Accompanied by FL?,fonttitle=\bfseries]
A common thread across all the scenarios discussed above is the geo-distributed nature of data generation. XR systems deployed in homes, hospitals, factories, classrooms, and entertainment spaces continuously collect embodied, multi-modal data, which is generated/collected locally on XR devices. Such a geo-distributed nature of data generation poses a major challenge to the integration of M3T FMs into XR scenarios: traditional \textit{centralized} M3T FM training paradigms require transferring/pooling data from all XR devices into a centralized server to train a globally shared model. Yet, in the context of XR, the centralization of XR data is impeded by three key limitations:
\vspace{-2mm}
\begin{itemize}[leftmargin=0.1mm, itemsep=-0.25em]
\item \textbf{Privacy:} XR data inherently includes deeply personal and behavioral information (e.g., emotional states, cognitive intentions, and biometric patterns) that are not suitable for off-device transmission or cloud storage.
\item \textbf{Bandwidth and Latency:} Continuously uploading high-dimensional sensory data from distributed XR devices imposes unsustainable demands on wireless infrastructure, introducing unacceptable delays in real-time experiences.
\item \textbf{Personalization:} Centralized training typically produces generic models that fail to reflect local variations in user behavior and context -- critical for applications such as neuroadaptive interfaces or emotion-aware storytelling.
\end{itemize}
\vspace{-2mm}

$\star$ Leveraging the distributed model training capabilities of FL, \textbf{M3T FedFMs} offer a compelling alternative that can address these challenges, making them a major component of next-generation intelligent XR systems.Specifically, as elaborated previously and illustrated in Fig.~\ref{fig:federated-learning}, integrating FL into M3T FedFMs enables on-device model training/fine-tuning, ensuring that user data remains local, either on the originating XR device or within the user's personal ecosystem of devices. This architecture prevents raw user data (e.g., emotional states, biometric signals) from being exposed to external entities as only model/gradients parameters are shared across the system. Moreover, by avoiding real-time data uploads and keeping voluminous sensory inputs local at user devices, this approach mitigates bandwidth and latency concerns. Finally, the inherent support for local model adaptation in FL allows M3T FedFMs to be tailored to individual users' behaviors and contexts, thereby offering model personalization.

\end{tcolorbox}

\underline{$\star$ In the following, as our focus is solely on `M3T FedFMs', we use a shorter abbreviation `FedFMs' to refer to them.}

\section*{Manifestations of \textbf{SHIFT} Dimensions: Why XR Represents a Unique Frontier for FedFMs?}
Although the integration of FL into FedFMs helps address the core challenges of FMs in XR ecosystems (namely, challenges related to \textit{`Privacy'}, \textit{`Bandwidth and Latency'}, and \textit{`Personalization'} as discussed above), the practical implementation of FedFMs in XR environments introduces a range of nuanced challenges. While several of these challenges, such as hardware heterogeneity, communication cost, and heterogeneous/non-IID data distribution, are shared with other FedFMs settings, their manifestation in XR environments is uniquely intensified by the real-time, immersive, and multi-modal nature of user interaction. In particular, XR devices must locally run large FMs while handling diverse multi-modal inputs, often in dynamic physical and virtual environments. These systems operate under strict constraints on power, compute, memory, storage, and thermal capacity, while simultaneously requiring low-latency, personalized responses to support interactive, task-varying, and context-aware user experiences. These challenges must be carefully characterized to inform future research directions, system design, and deployment strategies. Motivated by the need to isolate and understand the most relevant obstacles, we categorize the core challenges associated with training, adapting, and personalizing FedFMs in XR systems under \textbf{SHIFT} dimensions: (1) \underline{\textbf{S}}ensor and modality diversity, (2) \underline{\textbf{H}}ardware heterogeneity and system-level constraints, (3) \underline{\textbf{I}}nteractivity and embodied personalization, (4) \underline{\textbf{F}}unctional/task variability, and (5) \underline{\textbf{T}}emporality and environmental variability. It is worth mentioning that these challenges/dimensions are complementary to those of conventional FL (e.g., data heterogeneity~\cite{borazjani2025redefining}).

Below, we elaborate on each of \textbf{SHIFT} dimensions, highlighting how it imposes constraints that can disrupt standard FedFM learning assumptions and necessitate a rethinking of how FedFMs are designed, trained, and deployed in XR environments.

\subsubsection*{(Dimension 1) \underline{\textbf{S}}ensor and Modality Diversity}
XR devices may \textit{differ vastly} in the types and combinations of sensors they possess, ranging from lightweight AR glasses with vision-only input to immersive VR headsets equipped with eye-tracking, haptic feedback, spatial audio, and depth sensing. This sensor heterogeneity results in non-overlapping feature spaces across devices. Additionally, input modalities may \textit{dynamically appear or disappear} within an XR session: for instance, due to muted microphones, disabled haptics, or transient occlusions in the environment.
If left unaddressed, this heterogeneity and temporal variability in sensor modalities can severely degrade the performance and convergence of FedFMs. Specifically, devices with incomplete or shifting modality views may contribute noisy, biased, or incompatible model updates to the global model, leading to poor generalization, slower convergence, or even divergence during collaborative model training. Therefore, it is critical to design FedFMs that can robustly incorporate updates from devices with \textit{incomplete modality views} and \textit{temporally dynamic input streams}, ensuring that all participating devices, regardless of their sensing capabilities or environmental conditions, can contribute meaningfully to the collaborative model training process.

\begin{table*}[t]
\centering
\caption{Evaluation metrics for FedFMs in XR ecosystems. These metrics reflect both generic ML objectives and XR-specific requirements and are categorized by the relevant SHIFT dimensions.}
\vspace{-1.5mm}
\label{tab:ffm_xr_metricsFinal}
{\notsotiny
\renewcommand{\arraystretch}{1.5}
\begin{tabular}{| >{\centering\arraybackslash}m{1.9cm} 
|| >{\centering\arraybackslash}m{4.6cm} 
| >{\centering\arraybackslash}m{4.6cm}
| >{\centering\arraybackslash}m{4.6cm}|}
\toprule
\cellcolor{darkyellow}\textbf{Metric} & \cellcolor{darkyellow}\textbf{Description} & \cellcolor{darkyellow}\textbf{How to Measure} & \cellcolor{darkyellow}\textbf{SHIFT Dimension Relevance} \\
\midrule
\cellcolor{darkgreen}\textbf{Inference Latency} & \cellcolor{darkgreen}Time between the reception of sensing inputs and model output generation; critical for responsiveness. & \cellcolor{darkgreen} Log time between input event (e.g., gesture, gaze) and output generation. & \cellcolor{darkgreen} Real-time interaction \textbf{(I)} on diverse XR hardware \textbf{(H)} under varying conditions \textbf{(T)}. \\
\midrule
\cellcolor{lightgreen}\textbf{Personalization Effectiveness} & \cellcolor{lightgreen} Performance gain after local personalized (e.g., using prompt tuning) vs. generic global model. & \cellcolor{lightgreen} Compare task accuracy or user satisfaction rate before and after personalization. & \cellcolor{lightgreen} User-specific interaction \textbf{(I)} by adapting to diverse task requirements \textbf{(F)} in XR interfaces. \\
\midrule
\cellcolor{darkgreen}\textbf{Personalization Necessity Rate} & \cellcolor{darkgreen} Proportion of users or tasks where global FedFM meets threshold, avoiding need for adaptation. & \cellcolor{darkgreen} Count users/tasks for which the global model exceeds a baseline accuracy (e.g., 90\%) without fine-tuning. & \cellcolor{darkgreen}Balancing hardware use \textbf{(H)} and user needs \textbf{(I)} across tasks \textbf{(F)} in XR systems. \\
\midrule
\cellcolor{lightgreen}\textbf{Cross-Context Generalization} & \cellcolor{lightgreen} The model’s ability to generalize effectively across previously unseen users, environments, or tasks. & \cellcolor{lightgreen} Test model on data from new users or contexts not seen during training. & \cellcolor{lightgreen} Diverse sensor inputs \textbf{(S)}, tasks \textbf{(F)}, and dynamic XR environments \textbf{(T)}. \\
\midrule
\cellcolor{darkgreen}\textbf{Task-Specific Continuity} & \cellcolor{darkgreen} Retaining performance on previously learned tasks during adaptation on the new tasks. & \cellcolor{darkgreen} Measure accuracy drop on prior tasks after new-task fine-tuning (i.e., forgetting under continual learning). & \cellcolor{darkgreen} Performance consistency across evolving tasks \textbf{(F)} in dynamic XR settings \textbf{(T)}. \\
\midrule
\cellcolor{lightgreen}\textbf{Submodule Activation Efficiency} & \cellcolor{lightgreen} Average number of active FedFM modules per task; reflects model modularity. & \cellcolor{lightgreen} Count number of modules activated per task or per user interaction. & \cellcolor{lightgreen} Optimized resource use on varied XR hardware \textbf{(H)} for diverse tasks \textbf{(F)}. \\
\midrule
\cellcolor{darkgreen}\textbf{Model Storage Footprint} & \cellcolor{darkgreen} Total model size deployed on XR device, including adapters or quantized formats. & \cellcolor{darkgreen} Sum memory usage in MB for all active modules on device. & \cellcolor{darkgreen} Minimizing storage demands on resource-constrained XR hardware \textbf{(H)}. \\
\midrule
\cellcolor{lightgreen}\textbf{Communication Overhead} & \cellcolor{lightgreen} Amount of data transferred per aggregation round or XR session. & \cellcolor{lightgreen} Monitor the bits exchanged during global aggregation rounds or XR sessions. & \cellcolor{lightgreen} Data transfer for efficient operation on XR devices \textbf{(H)} in dynamic contexts \textbf{(T)}. \\
\midrule
\cellcolor{darkgreen}\textbf{Battery Impact / Energy Use} & \cellcolor{darkgreen} Device power consumption while running/adapting FedFM. & \cellcolor{darkgreen} Measure power draw in mW or energy used (Joules) via onboard or external sensors. & \cellcolor{darkgreen} Energy efficiency on XR hardware \textbf{(H)} for prolonged use in varied environments \textbf{(T)}. \\
\midrule
\cellcolor{lightgreen}\textbf{User-Reported QoE} & \cellcolor{lightgreen} Subjective user-perceived quality of experience, focusing on responsiveness and interaction fluidity. & \cellcolor{lightgreen} Collect Likert-scale ratings, qualitative feedback, or lag reports through user surveys. & \cellcolor{lightgreen} Immersive, responsive XR interactions \textbf{(I)} in dynamic settings \textbf{(T)}. \\
\midrule
\cellcolor{darkgreen}\textbf{Model-Inferred QoE} & \cellcolor{darkgreen} System-side estimation of perceived interaction quality without explicit user input. & \cellcolor{darkgreen} Monitor implicit behavioral signals like gaze jitter, task repetition, latency-induced hesitation, or input aborts. & \cellcolor{darkgreen} Seamless XR interaction \textbf{(I)} by detecting quality issues in real-time environments \textbf{(T)}.  \\
\midrule
\cellcolor{lightgreen}\textbf{On-Device Training Time} & \cellcolor{lightgreen} Duration required for local model adaptation, tuning, or fine-tuning directly on the XR device. & \cellcolor{lightgreen} Record duration of local fine-tuning routine on device per adaptation event. & \cellcolor{lightgreen} Speed of local adaptation on XR hardware \textbf{(H)} for personalized interactions \textbf{(I)} in dynamic contexts \textbf{(T)}. \\
\midrule
\cellcolor{darkgreen}\textbf{Privacy Leakage Risk} & \cellcolor{darkgreen} Potential data leakage via observation of model updates or representations during global aggregations. & \cellcolor{darkgreen} Run membership inference attacks, gradient inversion, or compute differential privacy bounds. & \cellcolor{darkgreen} Protecting sensitive sensor data \textbf{(S)} in XR ecosystems with diverse inputs. \\
\bottomrule
\end{tabular}
}
\end{table*}

\subsubsection*{(Dimension 2) \underline{\textbf{H}}ardware Heterogeneity and System-Level Constraints}
XR systems span a wide range of hardware from AR glasses with minimal compute and memory to high-end MR headsets equipped with advanced computing units. This hardware heterogeneity leads to \textit{asymmetric participation in model training}, where not all XR devices can host full models or conduct local model updates frequently. While conventional FL methods, and subsequently FedFMs, support asynchronous and partial device participation, the strict latency and reliability requirements of XR, such as sub-50ms response\footnote{Such latency requirements align with AR/VR studies \cite{di2019perceptual}, which show that delays over 50ms noticeably degrade system responsiveness and user comfort.} for gesture-based interaction, amplify the impact of hardware heterogeneity.

\subsubsection*{(Dimension 3) \underline{\textbf{I}}nteractivity and Embodied Personalization}
XR environments are not passive; instead, they are inherently interactive, responding to user's motion, attention, speech, and preference in real-time. This interactivity, coupled with the embodied nature of XR (where user movement, posture, and environment are integral to the interaction loop), introduces unique challenges for FedFM personalization \cite{hatami2024survey}. Specifically, FedFMs must maintain generalizability while enabling continuous, low-latency, on-device adaptation. In this context, the collaborative training paradigm of FL, as integrated into M3T FedFMs, allows devices to aggregate locally trained/fine-tuned model components, thereby enhancing model generalizability compared to isolated, non-collaborative training. Additionally, the previously described fine-tuning techniques employed in M3T FMs can potentially enable effective model personalization.
However, these fine-tuning strategies (e.g., prompt tuning, adding personalized task heads, or introducing adapters) may fall short in XR contexts, as they can be too resource-intensive or insufficiently responsive to support the real-time, low-latency demands of XR applications. Additionally, it remains unclear how frequently such local personalization should be triggered in practical XR scenarios, where user states and interactions evolve continuously in real-time.

\subsubsection*{(Dimension 4) \underline{\textbf{F}}unctional/Task Variability}
 In a single session/time-window of XR interactions, the user's tasks may vary from identifying objects and following spatial navigation cues, to engaging in conversations with a virtual assistant and responding to real-time gesture-based commands.
 These tasks differ not only in output format but also in temporal scale (e.g., real-time vs. near-real-time), relevant modality, and computational demand. This intra-user multi-tasking in XR is further compounded by \textit{inter-user and inter-application variability}: a user in a manufacturing setting may need real-time error detection and instructions \cite{mahjourian2024multimodal}, while a student in an educational XR system might engage in immersive simulations and voice-based tutoring \cite{yeganeh2025future}. This breadth of functionality, which we refer to as \textit{functional variability}, breaks the common assumption in conventional FedFMs that devices often share a fixed set of downstream tasks or training objectives.

\subsubsection*{(Dimension 5) \underline{\textbf{T}}emporality and Environmental Variability}
XR systems operate in environments that may change rapidly over time (e.g., users move through spaces with shifting lighting). Therefore, federated adaptations in FedFMs for XR must account for such rapid temporal variability while \textit{not} forgetting the previously learned knowledge, resembling the \textit{catastrophic forgetting} issue in conventional deep learning models \cite{kirkpatrick2017overcoming}. Nevertheless, addressing this challenge is underexplored in the literature on FedFMs.

Addressing the aforementioned \textbf{SHIFT} dimensions necessitates the development of novel techniques across training, fine-tuning, model storage/compression, and model communication within FedFM workflows. In the following, we discuss how such future techniques should be evaluated in terms of their performance. While this work does not aim to prescribe detailed technical research directions, we will later outline a set of high-level solution strategies that can serve as a basis for future exploration.

\begin{table*}[t]
\centering
\caption{Key design tradeoffs in deploying FedFMs in XR ecosystems. Each tradeoff highlights competing objectives that must be balanced to meet XR-specific constraints.}
\label{tab:ffm_tradeoffs}
\vspace{-1.5mm}
{\notsotiny
\renewcommand{\arraystretch}{1.5}
\begin{tabular}{| >{\centering\arraybackslash}m{3cm} 
|| >{\centering\arraybackslash}m{6.3cm} 
| >{\centering\arraybackslash}m{6.8cm}|}
\toprule
\cellcolor{darkyellow}\textbf{Design Tradeoff} & \cellcolor{darkyellow}\textbf{Options} & \cellcolor{darkyellow}\textbf{Why It Matters in XR} \\
\midrule

\cellcolor{lightgreen}\textbf{Local Personalization ~~~~~~~~~~~~~~~~~~~~~~~~~~~~~~~~~~~~~~~~~~~~~~~~~~~~~~~~~~ ~~~~~~~~~~~~~vs. ~~~~~~~~~~~~~~~~~~~~~~~~~~~~~~~~~~~~~~~~~~~~~~~~~~~~~~~~Global Generalization} & \cellcolor{lightgreen} 
\textbf{Option 1:} Fine-tune locally for user-specific behavior \newline
\textbf{Option 2:} Rely on a shared global FedFM & \cellcolor{lightgreen}
Personalization improves responsiveness and user alignment but increases on-device compute and energy consumption that may penalize the XR application. Global models without personalization may lack specificity for XR contexts. \\
\midrule

\cellcolor{darkgreen}\textbf{Update Frequency ~~~~~~~~~~~~~~~~~~~~~~~~~~ vs. ~~~~~~~~~~~~~~~~~~~~~~~~~~~~~~~~~~~~~~~~~~~~~~~~~~~~~~~~~~~~~~~~~Communication Overhead} & \cellcolor{darkgreen}
\textbf{Option 1:} Frequent federated updates for real-time adaptation \newline
\textbf{Option 2:} Sparse, event-triggered updates to save bandwidth & \cellcolor{darkgreen}
Frequent updates accommodate behavioral drift in the global model but increase on-device energy and bandwidth use that may penalize the XR application that runs on it. Sparse updates conserve resources but risk model obsolescence, degrading user experience over time. \\
\midrule

\cellcolor{lightgreen}\textbf{Centralized Aggregation ~~~~~~~~~~~~~~~~~~~~~~~~~~ vs. ~~~~~~~~~~~~~~~~~~~~~~~~~~ ~~~~~~~~~~~~~~~~~~~~~~~~~~ Decentralized D2D Aggregation} & \cellcolor{lightgreen}
\textbf{Option 1:} Aggregation via a central server (e.g., cloud-based aggregator) \newline
\textbf{Option 2:} D2D aggregation directly among nearby XR devices & \cellcolor{lightgreen}
Centralized aggregation ensures global model consistency and convergence, yet, it is dependent on cloud connectivity and may introduce latency. D2D aggregation allows opportunistic learning, but risks slow model convergence. 
\\
\midrule

\cellcolor{darkgreen}\textbf{User-Driven QoE Optimization ~~~~~~~~~~~~~~~~~~~~~~~~~~~~~~~~~~~~~~~~~~~~~~~~~~~~ vs. ~~~~~~~~~~~~~~~~~~~~~~~~~~~~~~~~~~~~~~~~~~~~~~~~~~~~ System-Driven QoE Optimization} & \cellcolor{darkgreen}
\textbf{Option 1:} Collect explicit feedback from users \newline
\textbf{Option 2:} Infer experience quality from behavioral signals & \cellcolor{darkgreen}
Explicit feedback is precise but intrusive; implicit inference is unobtrusive but error-prone. \\
\midrule

\cellcolor{lightgreen} \textbf{Modular Model Deployment ~~~~~~~~~~~~~~~~~~~~~~~~~~~~~~~~~~~~ vs. ~~~~~~~~~~~~~~~~~~~~~~~~~~~~~~~~~~~~ Monolithic Model Deployment} & \cellcolor{lightgreen}
\textbf{Option 1:} Use lightweight, shallow adapters/prompt-tuners for fast tuning \newline
\textbf{Option 2:} Fine-tune deeper layers of the local FM for higher fidelity & \cellcolor{lightgreen}
Shallow tuning conserves energy, while deep tuning (especially on rich data) enhances personalization at the cost of increased power and thermal stress. \\
\midrule

\cellcolor{darkgreen}\textbf{Privacy Preservation ~~~~~~~~~~~~~~~~~~~~~~~~~~vs. ~~~~~~~~~~~~~~~~~~~~~~~~~~~~~~~~~~~~~~~~~~~~~~~~~~~~Model Performance} & \cellcolor{darkgreen}
\textbf{Option 1:} Enforce strong privacy using differential privacy or encryption \newline
\textbf{Option 2:} Allow model updates and aggregation with no privacy protection  & \cellcolor{darkgreen}
Using privacy-protection methods guards XR users' confidentiality but may limit model quality and add resource overhead (e.g., when encryption methods are applied). Looser policies improve learning but raise concerns about data leakage. \\
\bottomrule
\end{tabular}
}
\end{table*}

\section*{Realization of Resource-Aware FedFMs for XR: Metrics, Tradeoffs, and Datasets}
Unlocking the potential of FedFMs in XR environments requires more than the development of novel algorithms; it demands a robust and interoperable ecosystem that reflects the unique challenges posed by the \textbf{SHIFT} dimensions. We offer our perspective on the various aspects of this ecosystem below. 

\textbf{FedFM Performance Evaluation in XR Ecosystems:}
To support the design and benchmarking of FedFMs in XR, we envision a targeted set of evaluation metrics that reflect the unique demands of XR systems, which are further aligned with the \textbf{SHIFT} dimensions. In Table~\ref{tab:ffm_xr_metricsFinal}, we provide a structured presentation of these metrics that can serve as a guide for developers and researchers to assess FedFM behavior across diverse XR settings. 

\textbf{Major FedFM Performance Tradeoffs in XR Ecosystems:}
Designing FedFMs for XR environments involves navigating a set of \textit{competing objectives/trade-offs} that emerge from the unique constraints of embodied interaction, mobile hardware, and real-time responsiveness. Subsequently, in Table~\ref{tab:ffm_tradeoffs}, we outline the key tradeoffs that researchers can consider when deploying FedFMs over XR devices.
It is worth mentioning that recent studies have started to explore such fundamental tradeoffs between computation, communication, and quality of experience for ML\cite{WangSCHS:23,wang2024bones} and other tasks\cite{chakareski2017vr,chakareski2022toward} in XR systems; however, extending them to FedFM settings remains an open direction.

\textbf{Datasets for FedFMs in XR:}  A core challenge in advancing FedFMs for XR is the lack of standardized datasets that reflect the embodied, multi-modal, and distributed nature of real-world XR systems. Table~\ref{tab:ffm_xr_datasets} summarizes both existing and our perspective on the needed/missing datasets, with a focus on their modality coverage, task diversity, and high-level description.
Our goal in presenting this table is twofold: (1) to identify available XR datasets that can be used for FedFM training and evaluation, and (2) to outline datasets that do not exist but are essential for benchmarking FedFMs under realistic XR conditions, helping future data collection efforts.

\begin{table*}[t]
\centering
\caption{Extended list of XR-relevant datasets for training and evaluating FedFMs. This table includes both real-world and needed datasets across multiple modalities and XR-specific challenges.}
\vspace{-1.5mm}
\label{tab:ffm_xr_datasets}
{\notsotiny
\renewcommand{\arraystretch}{1.5}
\begin{tabular}{| >{\columncolor{darkyellow}\centering\arraybackslash}m{2.5cm} 
|| >{\columncolor{darkyellow}\centering\arraybackslash}m{0.8cm} 
| >{\columncolor{darkyellow}\centering\arraybackslash}m{2.7cm} 
| >{\columncolor{darkyellow}\centering\arraybackslash}m{3cm}
| >{\columncolor{darkyellow}\centering\arraybackslash}m{6.3cm}|}
\toprule
\textbf{Dataset} & \textbf{Type} & \textbf{Modalities} & \textbf{Primary Tasks} & \textbf{Description} \\
\midrule

\rowcolor{lightgreen} \textbf{xR-EgoPose \cite{tome2019xr}} & Existing & Egocentric RGB & 3D human pose estimation & Head-mounted fisheye video dataset enabling full-body 3D pose recovery for XR applications. \\
\midrule

\rowcolor{darkgreen} \textbf{HOI4D \cite{liu2022hoi4d}} & Existing & RGB-D, 3D hand/object pose & Human-object interaction, affordance learning & Large-scale egocentric dataset with 2.4M frames covering daily indoor tasks and object manipulation. \\
\midrule

\rowcolor{lightgreen} \textbf{CWIPC-SXR \cite{reimat2021cwipc}} & Existing & RGB-D, point clouds, audio & Social interaction modeling, real-time rendering & Dynamic point cloud dataset capturing 45 sequences of human interactions in social XR scenarios; includes RGB-D, audio, etc. \\

\midrule

\rowcolor{darkgreen} \textbf{HEADSET \cite{ lohesara2023headset}} & Existing & RGB-D, point cloud, light field & Emotion recognition under occlusion & Volumetric capture of emotional facial expressions with head-mounted gear and occlusion scenarios. \\
\midrule

\rowcolor{lightgreen} \textbf{AMIS \cite{bhattacharya2025amis}} & Existing & Audio, video & Multimodal communication modeling & Open-source audiovisual dataset for analyzing immersive communication and social XR interaction. \\
\midrule

\rowcolor{darkgreen} \textbf{Federated XR Simulators} & \textbf{Needed} & Multimodal (vision, audio, haptics, speech, motion) & Multimodal fusion, personalized policy learning & Simulated XR environments (e.g., Unity/Omniverse) with structured profiles for device-user-task modeling in FL settings. \\
\midrule

\rowcolor{lightgreen} \textbf{Federated Role-Specific Benchmarks} & \textbf{Needed} & Multi-modal with user metadata & Role-conditioned intent prediction, task alignment & Pre-partitioned benchmarks reflecting user roles, hardware diversity, and task heterogeneity. \\
\midrule

\rowcolor{darkgreen} \textbf{Longitudinal XR Interaction Logs} & \textbf{Needed} & Vision, gaze, audio, task feedback, physiological signals & Continual learning, drift detection & Tracks evolving user behavior across time for learning drift, retention, and personalization analysis. \\
\midrule

\rowcolor{lightgreen} \textbf{Social and Collaborative XR Benchmarks} & \textbf{Needed} & Gaze, speech, multi-user context & Multi-agent coordination, shared task grounding & Captures collaborative and role-aware interactions in XR environments to train FedFM-based assistants. \\
\midrule

\rowcolor{darkgreen} \textbf{Cognitive Load-Aware Datasets} & \textbf{Needed} & Gaze, micro-pauses, EEG/fNIRS (optional) & Cognitive load estimation, pacing adjustment & Supports user state estimation via behavioral and physiological patterns for adaptive interface behavior. \\
\bottomrule
\end{tabular}
}
\end{table*}

\section*{Towards Resource-Aware and High-Performance FedFMs for XR under SHIFT Dimensions}

Effectively addressing the \textbf{SHIFT} dimensions requires a concrete multi-faceted research agenda tailored to the operational realities of XR environments, investigation of which deserves a separate study and is outside the scope of this perspective paper. Nevertheless, to stimulate further investigations in this untapped area we provide a set of \textit{high-level research directions} below.
\\

\noindent \textbf{\textit{(Solution 1) Addressing \underline{S}ensor and Modality Diversity:}} To address incomplete or intermittent modality signals, future works can explore the following directions:
\begin{itemize}[leftmargin=3mm, itemsep=-0.25em]
    \item \textbf{Robust Multimodal Fusion under Missing Inputs:} Develop FedFM models that can operate effectively with partial, missing, or dynamically changing sensory streams by leveraging redundancy across modalities and data augmentation\cite{alhoraibi2024generative}.
    \item \textbf{Asymmetric Input Adaptation:} Design FedFM training strategies where devices with fewer modalities (e.g., vision-only AR glasses) can improve the quality of their local FMs through knowledge sharing with rich-modality devices (e.g., VR headsets with gaze and audio) through techniques such as cross-modal distillation or embedding alignment.
\end{itemize}

\noindent \textbf{\textit{(Solution 2) Addressing  \underline{H}ardware Heterogeneity and System-Level Constraints:}} To address the broad range of compute, battery, and thermal limitations across XR devices, future works can explore the following directions:
\begin{itemize}[leftmargin=3mm, itemsep=-0.25em]
    \item \textbf{Resource-Aware Inference Paths:} Design model components (e.g., MoEs) that dynamically adjust execution based on real-time resource states (e.g., battery, compute).
   \item \textbf{Collaborative Cross-Device Training:} Enable extremely resource-limited devices (e.g., smart glasses) to securely offload their data to more capable co-owned devices (e.g., XR headsets, smartphones, or edge hubs), where the amount of shared data and device assignments are optimizable variables.   
   This paradigm leverages user-owned compute availability to maintain personalization while satisfying device constraints.
\end{itemize}

\noindent \textbf{\textit{(Solution 3) Addressing  \underline{I}nteractivity and Embodied Personalization:}}
To address the need for fast user-specific model adaptations in dynamic XR settings, future works can explore the following directions:
\begin{itemize}[leftmargin=3mm, itemsep=-0.25em]
    \item \textbf{Composable User-Specific Modules:} Design and optimize XR-specific plug-and-play components (e.g., adapters, task heads) that can quickly specialize to a user’s context without altering the shared global backbone. These modules could also be sourced from other users who have encountered similar data or interaction patterns, though identifying and transferring such modules across users remains an open challenge.
 \item \textbf{Behavior-Conditioned Personalization Signatures:} Develop low-overhead user ``signatures" (e.g., in the form of contextual prompts) derived from embodied interaction traces (e.g., motion rhythm, gaze dynamics, voice cadence) that are fed to the model and condition its responses without requiring full parameter updates, enabling personalization that is both efficient and privacy-preserving.
\end{itemize}

\noindent \textbf{\textit{(Solution 4) Addressing  \underline{F}unctional and Task Variability:}} To address the wide spectrum of tasks present in XR applications, future works can explore the following directions:
\begin{itemize}[leftmargin=3mm, itemsep=-0.25em]
    \item \textbf{Task-Isolated Update Paths:}  Implement task-specific modules (e.g., separate heads, adapter stacks, or routing gates) that can carefully isolate gradients during training, ensuring that updates from one task do not unintentionally degrade performance on others.
\item \textbf{Dynamic Task Module Selection:} Develop strategies for deciding whether to support a new task by extending an existing module, such as stacking an adapter on top of a previously trained task head, or by creating a new dedicated task head within the FedFM. These decisions should consider factors, including representational similarity between tasks, available device resources (e.g., memory, compute), and empirical indicators such as model convergence speed.
\end{itemize}

\noindent \textbf{\textit{(Solution 5) Addressing  \underline{T}emporality and Environmental Variability:}}
To address the environmental changes and shifts in user behavior and physical surroundings, 
future works can explore the following directions:

\begin{itemize}[leftmargin=3mm, itemsep=-0.25em]
     \item \textbf{Temporal Context Signaling Modules:} Introduce lightweight modules that explicitly encode temporal drift signals, such as a shift in gaze patterns, degraded gesture precision, or increased latency in user feedback, and feed them into the FedFM. These modules can act as control gates that adjust model behavior based on the real-time collected signals from the user and environment, while preventing the model from forgetting its past learned knowledge.
    \item \textbf{Continual Contextual Adaptation:} Apply continual learning algorithms on various FedFM modules (e.g., encoder, task heads, MoEs) that balance quick adaptation to new scenarios, while retaining prior context for long-term stability.
\end{itemize}

It is worth mentioning that following the development of each of the solutions, a comprehensive evaluation is essential: one that rigorously assesses performance using the previously defined metrics and design trade-offs.

\section*{Conclusion}
In this perspective paper, we highlighted the motivations for utilizing M3T FedFMs in XR ecosystems and the unique challenges of this vision framed by the \textbf{SHIFT} dimensions.
 We also provided a framework for performance evaluation to enable the resource-effective integration of M3T FedFMs into XR ecosystems. We finally envisioned a set of research directions, aiming to motivate further investigations in this underexplored area. 
We believe that challenges ahead are substantial, but so too are the potentials and benefits: intelligent XR systems that respect privacy, adapt seamlessly to users and environments, and generalize across a wide spectrum of devices, applications, and modalities.

\bibliography{BIB,ChakareskiBibliography}

\section*{Acknowledgements}
The authors acknowledge partial support from the National Science Foundation (NSF) under Grant No. 2421761. S. Hosseinalipour was supported by this grant during the course of this research.

\section*{Author contributions}
F. Nadimi and S. Hosseinalipour conceptualized the paper, prepared the first draft of the paper, and refined the paper.

\noindent  P. Abdisarabshali and K. Borazjani contributed to the ideas, editing the text, and drawing the diagrams. 

\noindent J. Chakareski contributed to the ideas and editing the text.

\noindent  The final version of the paper was critically
reviewed and approved by all authors.

\section*{Competing interests}
The authors declare no competing interests.

\section*{Ethical Considerations}
The pictures and diagrams that include human faces in this paper are digitally generated by graphic artists. They were not based on any real individuals and are not intended to depict, and do not knowingly depict, any identifiable person.

\section*{Additional information}
Correspondence and requests for materials should be addressed to
S. Hosseinalipour.
\end{document}